\documentclass[conference]{IEEEtran}
\IEEEoverridecommandlockouts
\usepackage{cite}
 \usepackage{amsmath,amssymb,amsfonts,amsthm}
 \usepackage{cases}
\usepackage{custom_macros}
\usepackage{graphicx}
\usepackage{textcomp}
\usepackage{xcolor}
\def\BibTeX{{\rm B\kern-.05em{\sc i\kern-.025em b}\kern-.08em
    T\kern-.1667em\lower.7ex\hbox{E}\kern-.125emX}}
   \usepackage{bbm} 
 \usepackage{multicol}
\usepackage[bookmarks=true]{hyperref}
\usepackage{xcolor}
\usepackage{tikz}
 \usepackage{tikz-3dplot}
 \usetikzlibrary{shapes.multipart}
 \usetikzlibrary{shapes.symbols}
\usepackage{pgfplots}

\usepackage{standalone}
\usepackage{cases}

\usepackage{subcaption}
\usepackage{comment}
\usepackage{etoolbox}
\usepackage{float}
\usepackage{algorithm}
\usepackage{algpseudocode}
\usepackage{dsfont}
\usepackage[belowskip=-15pt,aboveskip=0pt]{caption}

\allowdisplaybreaks
    \newcommand\Alpha{\mathrm{A}}
\newtheorem{lemma}{Lemma}

\newtheorem{theorem}{Theorem}

\newtheorem*{remarknn}{Remark}

\begin{document}
 
\title{Semi-Decentralized Federated  Learning \\ with Collaborative Relaying
\thanks{M. Yemini, R. Saha, and A.J. Goldsmith are partially supported by the AFOSR award \#002484665 and a Huawei Intelligent Spectrum grant.\\
E. Ozfatura and D. G\"{u}nd\"{u}z received funding from the European Research Council (ERC) through Starting Grant BEACON (no. 677854) and the UK EPSRC (grant no. EP/T023600/1) under the CHIST-ERA program.}
}

\author{\IEEEauthorblockN{Michal Yemini}
\IEEEauthorblockA{Princeton University }
\and
\IEEEauthorblockN{Rajarshi Saha}
\IEEEauthorblockA{Stanford University}
\and
\IEEEauthorblockN{Emre Ozfatura}
\IEEEauthorblockA{Imperial College London}
\and
\IEEEauthorblockN{Deniz G\"{u}nd\"{u}z}
\IEEEauthorblockA{Imperial College London}
\and
\IEEEauthorblockN{Andrea J. Goldsmith}
\IEEEauthorblockA{Princeton University}
}

\maketitle

\begin{abstract}
We present a semi-decentralized federated learning algorithm wherein clients collaborate by relaying their neighbors' local updates to a central parameter server (PS).
At every communication round to the PS, each client computes a local consensus of the updates from its neighboring clients and eventually transmits a weighted average of its own update and those of its neighbors to the PS.
We appropriately optimize these averaging weights to ensure that the global update at the PS  is unbiased and to reduce the variance of the global update at the PS, consequently improving the rate of convergence.
Numerical simulations substantiate our theoretical claims and demonstrate settings with intermittent connectivity between the clients and the PS, where our proposed algorithm shows an improved convergence rate  and accuracy in comparison with the federated averaging algorithm.
\end{abstract}

\section{Introduction}
\label{sec:introduction}
Federated learning (FL) algorithms iteratively optimize a common objective function to learn a shared model over data samples that are localized over multiple distributed clients \cite{FL1}. 
FL approaches aim to reduce the required communication overhead and improve clients' privacy by training local models of private dataset at the clients  and forwarding them periodically to a centralized parameter server (PS).
In practical FL setups, some clients are stragglers and cannot send their updates regularly, either because: \textit{(i)} they cannot finish their computation within a prescribed deadline, or \textit{(ii)} due to communication limitations \cite{chen2021distributed}, where they suffer from intermittent connectivity to the PS since their wireless channel is temporarily blocked \cite{6834753,8047278,yasamin,pappas,zavlanos,gilIJRR}.
Stragglers deteriorate the convergence of FL as the computed local updates become stale.
This can even result in bias in the final model in the case of persistent stragglers.
On the other hand, \textit{Communication stragglers} (type \textit{(ii)}) are inherently different from \textit{computation-limited stragglers} (type \textit{(i)}), since it can be solved by relaying the updates to the PS via neighboring clients.

Communication quality at the wireless edge as a key design principle is considered in the federated edge learning (FEEL) framework \cite{oac0}, which takes into account the wireless  channel characteristics from the clients to the PS to optimize the convergence and final model performance at the PS. 
So far the FEEL paradigm has mainly focused on  direct  communication  from the clients to the PS, and aimed at improving the performance by resource allocation across clients  \cite{FL.CS-RA1,FL.CS-RA2,FL.CS-RA3,FL.CS-RA4,oac0,oac1,oac2,oac3,oac_privacy1,oac_privacy2}; these approaches have ignored possible cooperation between clients in the case of intermittent communication blockages.

Motivated by our prior works \cite{yemini_et_al:globeom2020,yemini_et_al:TWC_cloud_cluster,HFL2}, where client cooperation is used to improve the connectivity to the cloud and to reduce the latency and scheduling overhead, this work 
proposes  a new FEEL paradigm, where the clients cooperate to mitigate the detrimental effects of communication stragglers. 
In our proposed method, clients share their local updates with neighbors so that each client sends to the PS a weighted average of its current update and those of its neighbors.
Using this approach, the PS receives new updates from disconnected clients, which would otherwise become stale and be discarded.
We demonstrate the success of our relaying scheme through both theoretical analysis and numerical simulations.

\paragraph*{Related Works}
Decentralized collaborative learning frameworks have been introduced as an alternative to centralized FL, in which the PS is removed to mitigate a potential communication bottleneck and a single point of failure \cite{decent1,decent2,decent3,decent4,decent5,decent6,decent7,decent8, decent9, decent_top1,decent_top2,decent_top3}.
In decentralized learning, each client shares its local model with its neighbors through device-to-device (D2D) communications, and model aggregation is executed at each client in parallel.
This aggregation strategy is determined at each client according to the network topology, i.e., the connection pattern between the clients.

An alternative approach to both centralized and decentralized schemes is  {\em hierarchical FL (HFL)}  \cite{HFL1,HFL2,HFL3,HFL4}, where multiple PSs are employed for the aggregation to prevent a communication bottleneck.
In HFL, clients are divided into clusters and a PS is assigned to each cluster to perform local aggregation. The aggregated models at the clusters are later aggregated at the main PS in a subsequent step to obtain the global model. 
HFL has significant advantages over centralized and decentralized schemes,  particularly when the communication takes place over wireless channels since it allows spatial reuse of available resources \cite{HFL2}. Nonetheless, HFL requires employing multiple PSs that may not be practical in certain scenarios.
Instead, the idea of hierarchical collaborative learning can be redesigned to combine hierarchical and decentralized learning, which is referred as  {\em semi-decentralized FL}, where the local consensus follows decentralized learning with D2D communications, whereas the global consensus is orchestrated by the PS \cite{semi-decent1,semi-decent2}.  One of the major challenges in FL that is not considered in \cite{semi-decent1,semi-decent2} is the partial client connectivity \cite{part_device1,part_device2}. 
Unequal client participation due to intermittent connectivity exacerbates the impact of data heterogeneity \cite{FL.noniid1,FL.noniid2,FL.noniid3,FL.noniid4}, and increases the generalization gap.

Most existing works on FL assume error-free rate-limited orthogonal communication links, with  an underlying communication protocol that takes care of wireless imperfections.
However, this separation between the communication medium and the learning protocol can be  suboptimal \cite{oac0}.  
An alternative approach treats the communication of the model updates to the PS as an uplink communication problem and jointly optimizes the learning algorithm and the communication scheme \cite{oac0}.
Within this framework is an original and promising approach known as  {\em over-the-air computation (OAC)} \cite{oac1,oac2,oac3}, which exploits the superposition property of wireless signals to convey the sum of the model updates that are transmitted by each client in an uncoded fashion. 
In addition to bandwidth efficiency, the OAC framework provides a certain level of anonymity to clients due to its superposition nature; hence, it can enhance the privacy of the participating clients \cite{oac_privacy1,oac_privacy2}.
We emphasize that in OAC, PS receives the aggregate model, and it is not possible to disentangle the individual model updates.
Therefore, any strategy that utilizes a PS side aggregation mechanism with individual model updates to address unequal client participation is not compatible with the OAC framework.
One of the major advantages of our proposed scheme is that it mitigates the drawbacks of unequal client participation without requiring the identity of transmitting clients or their individual updates at the PS.
Therefore, our solution is compatible with OAC.

Client connectivity is a particularly significant challenge in FEEL, where the clients and the PS communicate over unreliable wireless channels. 
Due to their different physical environments and distances to the PS, clients can have different connectivity to each other and the PS.
This problem has been recently addressed in \cite{FL.CS,FL.CS-RA1,FL.CS-RA2,FL.CS-RA3,FL.CS-RA4,FL.CS-RA5,FL.CS-RA6,FL.CS-RA7} by considering customized client selection mechanisms to  balance the participation of the clients and the latency for the model aggregation in order to speed up the learning process. 
We adopt a different approach to this problem, where instead of designing a client selection mechanism, or optimizing resource allocation to balance client participation, we introduce a {\em relaying} mechanism that takes into account the nature of individual clients' connectivity to the PS and ensures that, in case of poor connectivity, their local updates are conveyed to the PS with the help of their neighboring clients.

\paragraph*{Paper Organization}
Sec.~\ref{sec:system_model_for_collaborative_relaying} presents the FL system model and the proposed FL collaborative relaying scheme. 
Sec.~\ref{sec:convergence_analysis}  presents conditions for the unbiasedness of our proposed scheme and an analysis of the convergence rate.
Sec.~\ref{sec:optimizing_alphas} optimizes the convergence rate of our proposed scheme, while Sec.~\ref{sec:numerical_simulations}
presents numerical results that validate our theoretical analysis and highlight the performance improvement in terms of training accuracy. 
Finally, Sec.~\ref{sec:conclusions} concludes this paper.

\begin{remarknn}
Due to space limitations, all proofs are omitted, and can be found in an online extended version of this paper \cite{extended_ISIT_22_arXiv}.
\end{remarknn}

\section{System Model for Collaborative Relaying}
\label{sec:system_model_for_collaborative_relaying}
\begin{figure}[t]
\centering
\includegraphics[width = 0.7\columnwidth]{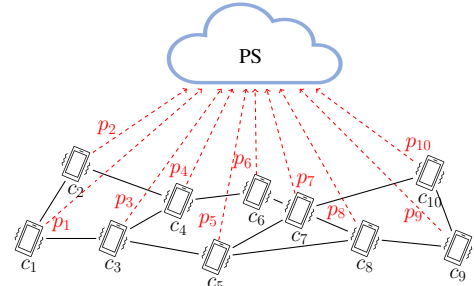}
\caption{System model with  intermittent uplink communication  between clients and PS (\textcolor{red}{dotted lines}) and reliable communication between neighboring clients (solid lines).}
\label{fig:communication_model}
\end{figure}
Consider $n$ clients communicating periodically with a PS that trains a global model $\xv \in \Real^d$.
Let $\Lc(\xv, \zetav)$ be the loss evaluated for a model $\xv$ at data point $\zetav$.
Denote the local loss at client $i$ by $f_i : \Real^d \times \Zc_i \to \Real$, where $f(\xv; \Zc_i) = \frac{1}{\lvert \Zc_i \rvert}\sum_{\zetav \in\Zc_i}\Lc(\xv; \zetav)$.
Here, $\Zc_i$ is the local dataset of client $i$.
The goal of PS is to solve the following empirical risk minimization (ERM) problem:\footnote{For simplicity, we assume $\lvert \Zc_i \rvert = \lvert \Zc_j \rvert$ for all $i,j \in [n]$. Our method can be extended to the setting of imbalanced local dataset sizes as well.}
\[
    \xv^* = \argminimize_{\xv \in \Real^d}f(\xv) \triangleq \argminimize_{\xv \in \Real^d} \frac{1}{n}\sum_{i=1}^{n}f_i(\xv;\Zc_i).
\]

\subsection{FL with Local SGD at Clients}
\label{subsec:FL_with_local_sgd_at_client}

Denote the local gradient as $\nabla f_i(\xv) \triangleq \nabla_{\xv} f(\xv;\Zc_i)$, and let $g_i(\xv)$ be an unbiased estimate of it.
In the $r^{th}$ \textit{round} of FL, the PS broadcasts the global model $\xv^{(r)}$ to the clients.
For a local averaging \textit{period} of $\Tc$, each client performs $\Tc$ iterations of local training, after which the local models are sent to the PS for aggregation.
For local iteration $k \in [0:\Tc]$ of the $r^{th}$ round, client $i$ applies the local update rule:
\begin{equation}
\label{eq:client_update_weighted_FL}
\xv_i^{(r,k+1)} = \xv_i^{(r,k)}-\eta_r\gv_i\left(\xv_i^{(r,k)}\right),
\end{equation}
where $\eta_r$ is the learning rate for round $r$ and $\xv_i^{(r,0)} = \xv^{(r)}$.

\subsection{Communication Model}
\label{subsec:communication_model}

\textbf{Communication between clients and PS}.
We consider a setting where the uplink connections between the clients and the PS are intermittent.
As shown in Fig.~\ref{fig:communication_model}, we model the connectivity of client $i$ to the PS at round $r$ by the Bernoulli random variable $\tau_i(r) \sim \text{Bern}(p_i)$, where $\tau_i = 1$ indicates the presence of an uplink communication opportunity, whereas $\tau_i(r) = 0$ indicates a blocked uplink.
For simplicity of exposition, we consider the  uplink channel to be either completely blocked or perfectly available without any noise, and the downlink from PS to clients does not suffer from intermittent dropouts.

\begin{remark}
\label{remark:known_comm_prob} 
The connectivity probabilities $\{p_i\}_{i \in [n]}$ can be easily estimated using pilot signals.
Moreover, clients can share their $p_i$ with each other using local links in a pre-training phase.
On the other hand, we do not assume that the instantaneous connectivity information, i.e., $\tau_i(r), r \in [n]$ is available to any of the clients.
\end{remark}

\textbf{Communication between clients}.
The connectivity between clients is modeled by an undirected graph $G = (V,E)$ where $V = [n]$ and $(i,j) \in E \iff$ client $i$ can communicate with client $j$.
Let $\Ncal_i \triangleq \{j\in V: \{i,j\}\in E\}$. We do not assume that the graph $G$ is connected. Instead, it can be composed of multiple connected subgraphs.

\subsection{Collaborative Relaying of Local Updates}
\label{subsec:collaborative_relaying_of_local_updates}

Let $\Delta\xv_i^{(r+1)}$ denote client $i$'s update at the end of $\Tc^{th}$ local iteration of round $r$, i.e., $\Delta\xv_j^{(r+1)}=\xv_j^{(r,\mathcal{T})}-\xv^{(r)}$.
We assume that client $i$'s update $\Delta\xv_j^{(r+1)}$ is readily available to its neighbors.
Then client $i$ computes a weighted average of its own update and those of its neighbors in $\Nc_i$, i.e.,
\begin{equation*}
    \Delta\widetilde{\xv}_i^{(r+1)}=   \hspace{-3mm}\sum_{j\in\mathcal{N}_i\cup\{i\}}\hspace{-3mm} \alpha_{ij}\Delta\xv_j^{(r+1)} = \hspace{-3mm}\sum_{j\in\mathcal{N}_i\cup\{i\}}\hspace{-3mm} \alpha_{ij}\left(\xv_j^{(r,\mathcal{T})}-\xv^{(r)}\right),
\end{equation*}
where $\alpha_{ij}$ is a non-negative importance weight assigned by client $i$ while relaying the client $j$'s update.
Note that weighted averaging entails a complexity of $O\left(\max_{i\in[n]}|\mathcal{N}_i|+1\right)$.

\subsection{PS Aggregation}
\label{subsec:PS_aggregation}

In our setting, the PS does not explicitly select the subset of clients from which it wants to receive information, rather it receives updates from all \textit{communicating} clients at the beginning of every round.
The PS uses the following re-scaled sum of received updates:
\begin{equation}
    \label{eq:averaging_update_weighted_FL}
    \xv^{(r+1)} = \xv^{(r)}+w\sum_{i\in[n]}\tau_i(r+1)
\Delta\widetilde{\xv}_i^{(r+1)}.
\end{equation}
This update can be computed over-the-air and does not require the PS to know the identities of the communicating clients.
We set $w\hspace{-1mm}=\hspace{-1mm}1/n$ to preserve the unbiasedness of the objective function at the PS, as discussed in Sec.~\ref{sec:convergence_analysis}.
Our Collaborative Relaying (ColRel) algorithm is presented in Algs.~\ref{algo:client_i_comp_forward} and \ref{algo:PS_comp_forward}.

\begin{algorithm}[t]
\caption{{\sc ColRel-Client}: Collaborative Relaying}
\label{algo:client_i_comp_forward}
{\bf Input:} Round index $r$, Step-size $\eta_r$, Local avg. period $\mathcal{T}$, Neighborhood of client $i$ $\mathcal{N}_i$, 
$\alpha_{ij}$ for every $j\in\mathcal{N}_i\cup\{i\}$.
{\bf Output}: $\Delta\widetilde{\xv}_i^{r+1}$.
\begin{algorithmic}[1]
\State Receive $\xv^{(r)}$ from  PS.
\State Set $\xv_i^{(r,0)}=\xv^{(r)}$.
\For{$k \gets 0$ \textbf{to} $\mathcal{T}-1$}

Compute (stochastic) gradient $g_i(\xv_i^{(r,k)}t)$.

$\xv_i^{(r,k+1)} = \xv_i^{(r,k)}-\eta_rg_i\left(\xv_i^{(r,k)}\right)$.
\EndFor
\State Set  $\Delta\xv_i^{r+1}=\xv_i^{(r,\mathcal{T})}-\xv^{(r)}$.
\State Send $\Delta\xv_i$ to every $j\in\mathcal{N}_i$.
\State Receive $\Delta\xv_j$ from every $j\in\mathcal{N}_i$.
\State Compute  $\Delta\widetilde{\xv}_i^{r+1}=\sum_{j\in\mathcal{N}_i\cup\{i\}} \alpha_{ij}\cdot\Delta\xv_j^{r+1}$.
\State Transmit (relay) $\Delta\widetilde{\xv}_i^{r+1}$ to the PS.
\end{algorithmic}
\end{algorithm}

\begin{algorithm}[t]
\caption{{\sc ColRel-PS}: PS Aggregation}
\label{algo:PS_comp_forward}
{\bf Input:} Number of rounds $R$,
a set of clients $[n]$.\\
{\bf Output}: Global model $\xv^{(R)}$.
\begin{algorithmic}[1]
\State Set $\xv^{(0)}=\boldsymbol{0}$
\For{$k \gets 0$ \textbf{to} $\mathcal{T}-1$}

Broadcast $\xv^{(r)}$ to all clients.

Set $\tau_i(r+1)=1 \text{ or } 0$ depending on connectivity.

Update $\xv^{(r+1)} = \xv^{(r)}+\frac{1}{n}\sum_{i\in[n]}\tau_i(r+1)
\Delta\widetilde{\xv}_i^{r+1}$
\EndFor
\end{algorithmic}
\end{algorithm}

\section{Convergence Analysis}
\label{sec:convergence_analysis}

\subsection{Unbiasedness of \textsc{ColRel} FL}
In our collaborative relaying scheme, the local update of a particular client $i$ can be transmitted to the PS by itself, or by one or more of its neighbors $j \in \Nc_i$.
Since the PS may be blind to the identities of the clients, the clients collaborate among themselves to ensure that this redundancy is mitigated.
This is done by appropriately choosing the weights $\alpha_{ij}$.
In particular, Lemma \ref{lemma:unbiased} gives a sufficient condition on the values of $\{\alpha_{ij}\}_{i,j \in [n]}$ that ensures that the aggregated global update at the PS is an unbiased estimate of $\frac{1}{n}\sum_{i \in [n]}\Delta\xv_i^{(r)}$, the true aggregate in the case of perfect channel connectivity.

\begin{lemma}
\label{lemma:unbiased}
Let $w = 1/n$ and $\{\alpha_{ij}\}_{i,j \in [n]}$ be such that
\begin{equation}
    \label{eq:alpha_sum_unbiased_constraint}
    \mathbb{E}\left[\sum_{j\in\mathcal{N}_i\cup\{i\}} \hspace{-2mm}\tau_j(r+1)\alpha_{ji}\right] 
=p_i\alpha_{ii}+\sum_{j\in\mathcal{N}_i}p_j\alpha_{ji}= 1.
\end{equation}
Then, for every $i \in [n]$,
\begin{equation*}
    w\cdot\mathbb{E}\left[\sum_{j\in\mathcal{N}_i\cup\{i\}} \hspace{-2mm}\tau_j(r+1)\alpha_{ji}\Delta\xv_i^{r+1}\Big|\Delta\xv_i^{r+1}\right] =\frac{1}{n}
\Delta\xv_i^{r+1}.
\end{equation*}
\end{lemma}
Note that the standard FL setting under intermittent client connectivity to the PS but with no collaboration between the clients is captured by the choice $w = 1/n$, $\Nc_i = \emptyset, p_i = p, \alpha_{ii} = 1, \alpha_{ij} = 0$ for all $i, j \in [n]$ and $j \neq i$.

\subsection{Expected Suboptimality Gap}
Next,   Thm.~\ref{theorem:expected_distance_to_opt} presents the convergence rate of \textsc{ColRel} as a function of $\{\alpha_{ij}\}$, under the following assumptions.
\begin{assumption}
\label{assumption:Lipschitz}
For any $i$, the local loss $f_i$ is $L$-smooth w.r.t. $\xv$, i.e., for any $\xv, \yv \in \Real^d$, $\lVert \nabla f_i(\xv) - \nabla f_i (\yv) \rVert_2 \leq L \lVert \xv -\yv \rVert_2$.
\end{assumption}
\begin{assumption}
\label{assumption:unbiased_bounded_stochastic_gradients}
The stochastic gradients $\gv_i(\xv)$ are unbiased and have bounded variance, i.e.,  $\forall\: i \in [n]$:\\
1) $\mathbb{E}[\gv_i(\xv)] = \nabla f_i(\xv)$, and \\
2) $\mathbb{E}\lVert \gv_i(\xv) - \nabla f_i(\xv) \rVert_2^2 \leq \sigma^2$ for some finite $\sigma^2$.
\end{assumption}
\begin{assumption}
\label{assumption:Mu_stongly_convex}
For any $i$, the loss $f_i$ is $\mu$-strongly convex, i.e., for any $\xv, \yv \in \Real^d$, $(\nabla f_i(\xv)- \nabla f_i(\yv))^\top(\xv - \yv) \geq \mu\lVert \xv -\yv \rVert_2^2$.
\end{assumption}

Let $\Av = (\alpha_{ij})_{i,j \in [n]}$ denote the $n \times n$ matrix of relay weights, and let $\Nc_{il} = (\Nc_i \cup \{i\})\cap (\Nc_l \cup \{l\})$ denote the common neighborhood of nodes $i$ and $j$.
Suppose,
\begin{equation}
    S(\boldsymbol{p},\boldsymbol\Alpha)=\sum_{i,l\in[n]}\sum_{j: j\in\mathcal{N}_{il}} p_j(1-p_j)\alpha_{ji}\alpha_{jl}.
\end{equation}

\begin{theorem}
\label{theorem:expected_distance_to_opt}
Under Asms.~\ref{assumption:Lipschitz}-\ref{assumption:Mu_stongly_convex} and condition \eqref{eq:alpha_sum_unbiased_constraint}, \textsc{ColRel}, as specified by Algs.~\ref{algo:client_i_comp_forward} and \ref{algo:PS_comp_forward},  with $\eta_r=\frac{4\mu^{-1}}{r\mathcal{T}+1}$, satisfies for every $r \geq r_0(\boldsymbol{p},\boldsymbol\Alpha)$,
\begin{align*}
\mathbb{E}\lVert\xv^{(r+1)}-x^{\star}\rVert^2 \leq &\frac{(r_0\mathcal{T}+1)}{(r\mathcal{T}+1)^2}\|\xv^{(0)}-x^{\star}\|^2 + \frac{C_1(\boldsymbol{p},\boldsymbol\Alpha)\mathcal{T}}{k\mathcal{T}+1} \\ + &C_2\frac{(\mathcal{T}-1)^2}{k\mathcal{T}+1}+C_3(\boldsymbol{p},\boldsymbol\Alpha)\frac{\mathcal{T}}{(k\mathcal{T}+1)^2},
\end{align*}
where $B(\boldsymbol{p},\boldsymbol\Alpha) =\frac{2L^2}{n^2}S(\boldsymbol{p},\boldsymbol\Alpha)$, $ C_1(\boldsymbol{p},\boldsymbol\Alpha) = \frac{4^2}{\mu^2}\cdot \frac{2\sigma^2}{n^2}S(\boldsymbol{p},\boldsymbol\Alpha)$,
$C_2 = \frac{4^2}{\mu^2}\cdot L^2\frac{\sigma^2}{n}e$,
$C_3(\boldsymbol{p},\boldsymbol\Alpha) = \frac{4^4}{\mu^4}\cdot\left( L^2\sigma^2e+\frac{2L^2\sigma^2e}{n^2}S(\boldsymbol{p},\boldsymbol\Alpha)\right)$,  and $r_0(\boldsymbol{p},\boldsymbol\Alpha)= \max\left\{\frac{L}{\mu},4\left(\frac{B(\boldsymbol{p},\boldsymbol\Alpha) }{\mu^2}+1\right),\frac{1}{\mathcal{T}},\frac{4n}{\mu^2\mathcal{T}}\right\}$.
\end{theorem}
As a consequence of Thm.~\ref{theorem:expected_distance_to_opt}, it follows that,
\begin{equation}
    \mathbb{E}\left\|\xv^{(r+1)}-x^{\star}\right\|^2=O\left(\frac{\left\|\xv^{(0)}-x^{\star}\right\|^2}{r^2}+\frac{S(\boldsymbol{p},\boldsymbol\Alpha)}{r}\right).
\end{equation}
Therefore, the convergence rate can be improved by minimizing the term $S(\pv,\Av)$ subject to the unbiasedness condition in Lemma \ref{lemma:unbiased}.
Minimizing $S(\boldsymbol{p},\boldsymbol\Alpha)$ can also reduce $r_0(\boldsymbol{p},\boldsymbol\Alpha)$.

\section{Optimizing the relaying weights}
\label{sec:optimizing_alphas}

\begin{algorithm}[t]
\caption{{\sc OPT-$\alpha$:} Optimization of relay weight matrix $\boldsymbol{A}$}
\label{algo:opt_alpha_centralized}
{\bf Input:} Connectivity graph $G$, Transmission probability vector $\pv$, Maximum number of iterations $L$.\\
{\bf Output}: Relay weight matrix $\Av^{(L)}$ that approximately solves \eqref{eq:minimize_distance}.
\begin{algorithmic}[1]
\State Set $\boldsymbol\Alpha_{ji}^{(0)}=\frac{1}{(|\mathcal{N}_i|+1)\cdot p_{j}}\cdot\mathds{1}_{\{j\in\mathcal{N}_i\cup\{i\}:p_j>0\}}$.
\For{$\ell \gets 0$ \textbf{to} $L-1$}

Set $\ell \gets \ell + 1$.

Set $i=\ell\mod{n}+n\cdot\mathds{1}_{\{\ell\mod{n}=0\}}$.

Compute $\widehat{\boldsymbol\Alpha}_i^{(\ell)}$ according to \eqref{eq:alpha_i_clac_cur}.

Set $\boldsymbol\Alpha_k^{(\ell)}$ according to \eqref{eq:A_hat_iteration_ell}  for every $k\in[n]$.
\EndFor
\end{algorithmic}
\end{algorithm}

We choose the relay weight matrix $\Av$ to minimize the upper bound on the expected distance to optimality as given by Thm.~\ref{theorem:expected_distance_to_opt}.
Thus, we solve the following optimization problem:
\begin{align}
    \label{eq:minimize_distance}
    &\arg\min_{\boldsymbol\Alpha} S(\boldsymbol{p},\boldsymbol\Alpha),\nonumber\\
    &\text{s.t.:}\sum_{j: j\in\mathcal{N}_{i}} p_j\alpha_{ji}=1,\quad   \alpha_{ji}\geq 0\quad \forall i,j\in[n].
\end{align}
The function $S(\pv, \Av)$ is convex with respect to (w.r.t.) $\Av$ for $\pv \in [0,1]^n$.
It can be shown that the domain of \eqref{eq:minimize_distance} is separable w.r.t. $\Av_i$, the $i^{th}$ column of $\Av$, and we can use the Gauss-Seidel method \cite[Prop. 2.7.1]{Bertsekas_nonlinear_programming} to iteratively solve \eqref{eq:minimize_distance}.
At every iteration $\ell$, we compute the estimate $\Av^{\ell}$ as

\begin{flalign}
\label{eq:A_hat_iteration_ell}
\boldsymbol\Alpha_i^{(\ell)}=\begin{cases}
\widehat{\boldsymbol\Alpha}_i^{(\ell)} & \text{if } \ell\hspace{-0.25cm}\mod{n}+n\cdot\mathds{1}_{\{\ell\hspace{-0.25cm}\mod{n}=0\}}=i,\\
\boldsymbol\Alpha_i^{(\ell-1)}& \text{\:\hspace{-0.5mm}otherwise}
\end{cases}.
\end{flalign}

Here, $\widehat{\boldsymbol\Alpha}_{i}^{(\ell)}$ is given by
\begin{flalign}
\label{eq:opt_alpha_allocation_iterative}
\widehat{\boldsymbol\Alpha}_{i}^{(\ell)}&=\arg\min\sum_{j\in\mathcal{N}_{i}\cup\{i\}} p_j(1-p_j)\alpha_{ji}^2 \nonumber \\ 
&+ 2\sum_{l\in[n],l\neq i}\sum_{j\in\mathcal{N}_{il}} p_j(1-p_j)\alpha_{ji}\alpha_{jl}^{(\ell-1)},\nonumber\\
&\text{s.t.:}\sum_{j\in\mathcal{N}_{i}\cup\{i\}} p_j\alpha_{ji}=1,\quad  \alpha_{ji}\geq 0\quad \forall j\in[n].
\end{flalign}
Let $L_{ji}=\{l:l\in[n],l\neq i,j\in\mathcal{N}_{il}\}$, that is, the set of all clients that have $j$ as a mutual neighbor with $i$, and let $\beta_{ji}=\sum_{l\in L_{ji}} \alpha_{jl}^{(\ell-1)}$.
Let $\overline{p}(i) = \max_{k \in \Nc_i \cup\{i\}} p_k$.
Using Lagrange multipliers we solve  \eqref{eq:opt_alpha_allocation_iterative} for $j \in \Nc_i \cup \{i\}$ as follows:
\begin{flalign}
\label{eq:alpha_i_clac_cur}
\hspace{-0.2cm}\widehat{\alpha}_{ji}^{(\ell)} = 
\begin{cases}
\left(-\beta_{ji}+\frac{\lambda_i}{2(1-p_j)}\right)^+&
\hspace{-2mm}\text{if } p_j\in(0,1), \overline{p}(i) < 1,\\
\frac{1}{\sum_{k\in[n]}\mathds{1}_{\{p_k=1,k\in\mathcal{N}_i\cup\{i\}\}}} & \hspace{-2mm}\text{if } p_j=1,\\
0 & \hspace{-2mm}\text{otherwise}. 
\end{cases}
\end{flalign}
Here, $\lambda_i$ satisfies $\sum_{j\in\mathcal{N}_{i}\cup\{i\}}p_j\left(-\beta_{ji}+\frac{\lambda_i}{2(1-p_j)}\right)^+=1$, and $(\cdot)^+ \triangleq \max\{\cdot, 0\}$. We can find $\lambda_i$ using the bisection method.
The complete algorithm is detailed in Alg.~\ref{algo:opt_alpha_centralized}.
Its overall \textit{complexity} is $O(L\cdot( n^2+K))$, where $K$ is the number of iterations used in the bisection method for optimizing $\lambda_i$.
\begin{remark}
The optimization \eqref{eq:minimize_distance} only requires  client $i$ to know the weight values for its neighbors of distance 2. Thus, we can exploit the communication links between clients, and optimize \eqref{eq:minimize_distance} distributively. We present the distributed algorithm in \cite{extended_ISIT_22_arXiv}.  
\end{remark}

\section{Numerical Simulations}
\label{sec:numerical_simulations}

We consider training a ResNet-$20$ model for image classification on CIFAR-$10$ dataset over $10$ clients; each executes $8$ local training steps of local-SGD.
All plots have been averaged over $5$ different realizations. 
We used a learning rate of $0.1$ for SGD, a coefficient of $1e-4$ for $\ell_2$-regularization to prevent overfitting, and a batch-size of $64$. 

\begin{figure}[t]
\centering
  \includegraphics[width=\linewidth,trim={1cm 2.3cm 1.5cm 3.8cm},clip]{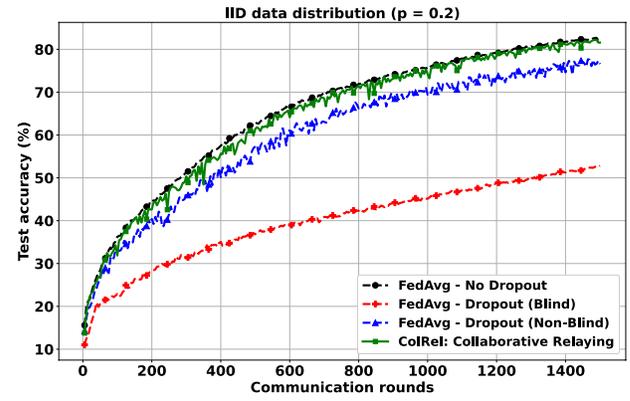}
  \caption{Homogeneous connectivity with $p_i=0.2,\forall i\in[n]$ and FCT.}
  \label{fig:iid_dropout_pt2}
  \vspace{-0.45cm}
\end{figure}%

\begin{figure}[t] 
\centering
  \includegraphics[width=\linewidth,trim={1cm 2.3cm 1.5cm 3.8cm},clip]{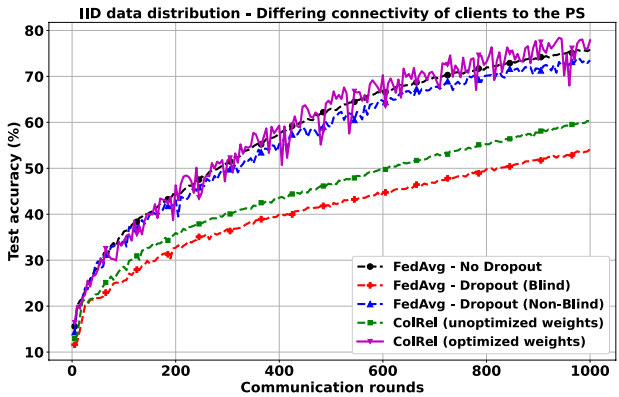}
  \caption{Different connectivity across clients with a ring topology.}
  \label{fig:iid_dropout_diffp_Ring}
\end{figure}%
\begin{figure}[t] 
\centering
  \includegraphics[width=\linewidth,trim={1cm 2.3cm 1.5cm 3cm},clip]{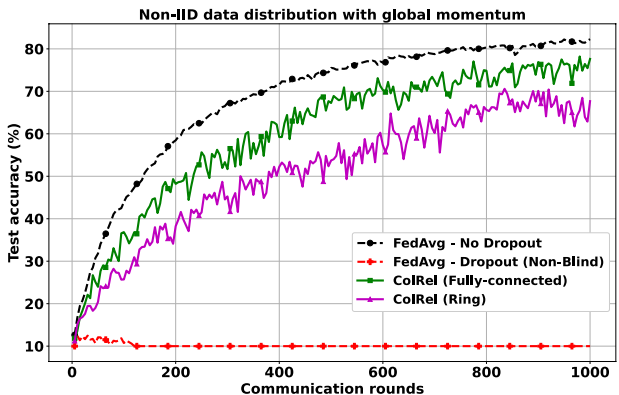}
  \caption{Non-IID data + global momentum.}
  \label{fig:noniid_global_momentum}
\end{figure}%

In Figs.~\ref{fig:iid_dropout_pt2} and \ref{fig:iid_dropout_diffp_Ring}, the dataset is distributed across the clients in an IID fashion.
As benchmarks, we consider \textit{Federated Averaging (FedAvg) - No Dropout}, in which all clients are able to successfully transmit their local updates to the PS at every communication round.
We also consider \textit{FedAvg - Dropout}, in which the PS is unaware of the identity of the clients, and simply assumes that the update for any client unable to successfully transmit is zero.
These benchmarks serve as natural upper and lower bounds to the performance of the proposed algorithm.

In Fig.~\ref{fig:iid_dropout_pt2}, we have a homogeneous connectivity setup with equal probability $p_i = 0.2$ that client $i\in[n]$ successfully transmits its local updates to the PS. Furthermore, we assume a fully-connected topology (FCT) where each clients is connected to all the other clients in the system.
\textsc{ColRel} achieves a performance on par with \textit{FedAvg - No Dropout}.
We also consider a non-blind strategy, \textit{FedAvg - Dropout (Non-Blind)} where the PS is aware of the identity of the clients, and knows exactly which clients have been successful in sending their local updates to the PS.
This is common in point-to-point learning settings.
In this case the PS simply ignores the clients that have been unable to send their updates, and averages the successful updates by dividing the global aggregate at the PS by the number of successful transmissions.

In Fig.~\ref{fig:iid_dropout_diffp_Ring} (and also in Fig.~\ref{fig:noniid_global_momentum}), we consider every client has a different probability of successful transmission to the PS according to $\pv = [0.1, 0.2, 0.3, 0.1, 0.1, 0.5, 0.8, 0.1, 0.2, 0.9]$.
We have deliberately chosen some clients to have a very low connectivity, some others moderate, and others very high.
We consider a ring topology where client $i$ is connected to clients $(i-1)\mod{n}$ and $(i+1)\mod{n}$.
For this setting, we distinguish the cases with and without optimized weights.
The weights are optimized in order to minimize the term $S(\mathbf{p}, \mathbf{A})$, which consequently minimizes the variance of the iterates, subject to ensuring that the updates are unbiased according to Alg.~\ref{algo:opt_alpha_centralized}.
Note that explicitly optimizing the consensus weights that the clients use for their neighbors was not essential in Fig. \ref{fig:iid_dropout_pt2} because the initial weights of Alg.~\ref{algo:opt_alpha_centralized} are optimal  for a FCT with homogeneous connectivity to the PS, i.e., $p_i=p\:\forall i\in[n]$.

Finally, in Fig.~\ref{fig:noniid_global_momentum}, we consider the setting in which the training data is distributed across the clients in a non-IID fashion.
To emulate non-IID-ness, we consider the {\em sort-and-partition} approach in which the training data is initially sorted based on labels, and then divided into blocks and distributed among clients in a skewed fashion so that each client has data from only a few classes.
For the ring topology in this plot, we have considered each client to be connected to $4$ of its nearest neighbors.
We also use global momentum at the PS to update the global model.
Remarkably, FedAvg (even with non-blind averaging) fails to converge in this setting.
This is because in the absence of collaboration,  clients that  have important training samples that are critical for training a good model with high accuracy, may have a low probability of successful transmission and thus are rarely able to convey their updates to the PS. 
Therefore, when these clients are unable to convey their updates to the PS, the resulting test accuracy of the global model is $\sim10\%$, as good as a random classifier for $10$ classes.
Collaborative relaying ensures that the information from these critical datapoints are also conveyed to the PS even when the data owner does not have connectivity to the PS.

\section{Conclusions}
\label{sec:conclusions}

Our goal in this paper is to mitigate the detrimental effect of clients'  intermittent connectivity  on the training accuracy of FL systems.
For this purpose, we proposed a collaborative relaying strategy, which exploits the connections between clients to relay potentially missing model updates to the PS due to blocked clients. 
Our  algorithm allows the PS to receive an unbiased estimate of the model update, which would not be possible without relaying. 
We optimized the consensus weights at each client to improve the rate of convergence. 
Our proposed approach can be implemented even when the PS is blind to the identities of clients which successfully communicate with it at each round. 
Numerical results showed the improvement in training accuracy and convergence time that our approach provides under various settings, including IID and non-IID data distributions, different communication graph topologies, as well as blind and non-blind PSs.

\bibliographystyle{IEEEtran}

\end{document}